\documentclass[10pt,twocolumn]{article} 
\usepackage{simpleConference}
\usepackage{times}
\usepackage{graphicx}
\usepackage{amssymb}
\usepackage{url,hyperref}

\begin{document}

\title{AinnoSeg: Panoramic Segmentation with High Perfomance}

\author{Jiahong Wu$^{1}$,Jianfei Lu$^{2}$,Xinxin Kang$^{2}$,Yiming Zhang$^{3}$,Yinhang Tang$^{1}$,Jianfei Song$^{1}$,Ze Huang$^{1}$,\\
Shenglan Ben$^{1}$,Jiashui Huang$^{1}$,Faen Zhang$^{1}$\\
\\
{$^1$AInnovation Technology Co. Ltd,$^2$Peking University,$^3$University of Washington} \\
\{wujiahong,tangyinhang,songjianfei,huangze,benshenglan,huangjiashui,zhangfaen\}@ainnovation.com, \\
\{lujianfei,kangxinxin\}@pku.edu.cn ,\{ymz15\}@uw.edu\\
\\
}

\maketitle
\thispagestyle{empty}

\begin{abstract}
Panoramic segmentation is a scene where image segmentation tasks is more difficult. With the development of CNN networks, panoramic segmentation tasks have been sufficiently developed.However, the current panoramic segmentation algorithms are more concerned with context semantics, but the details of image are not processed enough. Moreover, they cannot solve the problems which contains the accuracy of occluded object segmentation,little object segmentation,boundary pixel in object segmentation etc. Aiming to address these issues, this paper presents some useful tricks. (a) By changing the basic segmentation model, the model can take into account the large objects and the boundary pixel classification of image details. (b) Modify the loss function so that it can take into account the boundary pixels of multiple objects in the image. (c) Use a semi-supervised approach to regain control of the training process. (d) Using multi-scale training and reasoning. All these operations named AinnoSeg, AinnoSeg can achieve state-of-art performance on the well-known dataset ADE20K \cite{zhou2017scene}
\end{abstract}

\section{Introduction}
Panoramic segmentation is an image segmentation task proposed in recent years. The main task is to segment all objects with regular attributes in the image. The panoramic segmentation task needs to segment all the regular objects in the image, so it can be used in the field of industrial vision,autonomous driving,retail and other fields. The existing panoramic segmentation algorithm has evolved from the fully convolutional networks\cite{long2015fully}. After experiencing the emergence of DeepLab v1, v2, v3 \cite{chen2017deeplab}, and the attention mechanism\cite{fu2019dual}, etc., more attention is paid to the context of the image. HRNet\cite{wang2020deep} begins to pay more attention to large-scale images from the upper layers Features. OCRNet\cite{yuan2019object} focus on the relationship between image objects and its pixels. However, these algorithms can't handle the relationship between each object in the image and the pixels it owns, and the detail boundary pixels of the object cannot be processed well.
\\
Most of the existing methods are mainly to build larger image receptive fields and texture information by lifting the pixels of the feature map, such as HRNet\cite{wang2020deep},PSPnet\cite{zhao2017pyramid}, and improve the relationship between the pixels in the feature image by applying the attention mechanism to define the category of pixels, Such as DANe\cite{fu2019dual}. Other methods use the relationship between pixels and objects to determine the type of pixels, such as OCRnet\cite{yuan2019object}. These methods only improve the ability of image segmentation to a certain extent. None of them solved the problem of large object boundary pixel segmentation and small object segmentation in the panoramic segmentation task.\\

In this work, in order to make the model have a better segmentation performance on the boundary pixels of large objects and have the ability for segmenting the little objects, we changed the basic network model so that it can pay more attention to the classification of the boundary pixels of large objects. At the same time, we use some useful tricks to improve the ability to segment small objects. Our main contributions can be summarized as three folds. 1). We change the basic model so that it can pay more attention to the boundary pixel values of large objects. 2). Some data augmentation methods are used to improve small object segmentation. 3). Semi-supervised methods are applied to create coarse-grained label and we put them into the training model to continuously improve the model performance. 4). We use multi-scale training and inferencing strategy to get the state-of-art performance on the famous dataset ADE20K.
  
\section{Related Work}
\subsection{Traditional Method}
Threshold segmentation\cite{otsu1979threshold} is the simplest method of image segmentation and also one of the most common parallel segmentation methods. It directly divides the image gray scale information processing based on the gray value of different targets. Edge Detection Segmentation algorithm\cite{yuheng2017image} uses Sobel operator and Laplacian Operator, and clustering K-means algorithm\cite{kanungo2002efficient} is also applied. The advantages of these traditional algorithms are simple calculation and faster operation speed. The disadvantage is that it can not achieve a good image segmentation performance.

\subsection{Deep Learning Method}

\textbf{The Method Of Increasing The Receptive Field:} From the early FCN\cite{long2015fully}, the upsampling method was used to increase the receptive field of feature map, then the PFPN\cite{kirillov2019panoptic} method was added later to increase the receptive field of image in the form of the feature pyramid. Dense-ASPP\cite{yang2018denseaspp} uses atrous
convolutions to build the Atrous Spatial Pyramid Pooling to increase the receptive field of the image. SegNet\cite{badrinarayanan2017segnet} uses an encoder-decoder model and retains  high-quality feature maps. The recently proposed HRNet\cite{wang2020deep} can keep the specific features of the low-level image by connecting all levels of the network, while ensuring high-level semantic features and expanding the receptive field of the segmentation model. Although these methods can alleviate some of the problems of segmentation accuracy, but they cannot solve the related boundary pixel relationship with the objects in the image, and also they cannot focus on segmentation of small objects.

\textbf{Pixel Association Method:} A major difficulty of panoramic segmentation is that regular objects must be classified in the image. There are so many different categories to arrange objects with wide variety of shapes and complex poses. To solve this problem, some algorithms use the attention mechanism. By associating each pixel of the feature map with the global pixel value, the classification effect of each pixel is obtained. The typical algorithm is DANet\cite{fu2019dual}, and there is also a conditional random field\cite{sutton2006introduction}. In this way, each pixel is related to the pixel at its boundary. The CRF algorithm applies the voting mechanism to use around pixels. ACNet\cite{ding2019acnet} and OCRNet\cite{yuan2019object} use the internal pixel values of each object in the image to determine the classification of boundary pixels. These methods can help get better segmentation performance on the boundary pixels, but they are still not effective enough. When the boundary of the object is not clear, the object in the image is blocked.

\section{Method}
Our improvement method includes: we use the basic HRNet\cite{wang2020deep} as backbone, and add Nonlocal\cite{wang2018non} and DANet\cite{fu2019dual} attention mechanism to make the model pay more attention to the boundary pixel relationship with its object. Then we use data augmentation methods like Mosaic\cite{bochkovskiy2020yolov4} to make the model pay more attention to the image segmentation effect of small objects. In addition, we use semi-supervised\cite{chen2020leveraging} form to create coarse-grained data and label,  and incorporate it into the model for auxiliary training. Finally  multi-scale training and inference are applied to achieve state-of-art performance on the famous panoramic segmentation dataset ADE20K\cite{zhou2017scene}.

\subsection{Network}
\begin{figure}[t]
\includegraphics[width=1\linewidth]{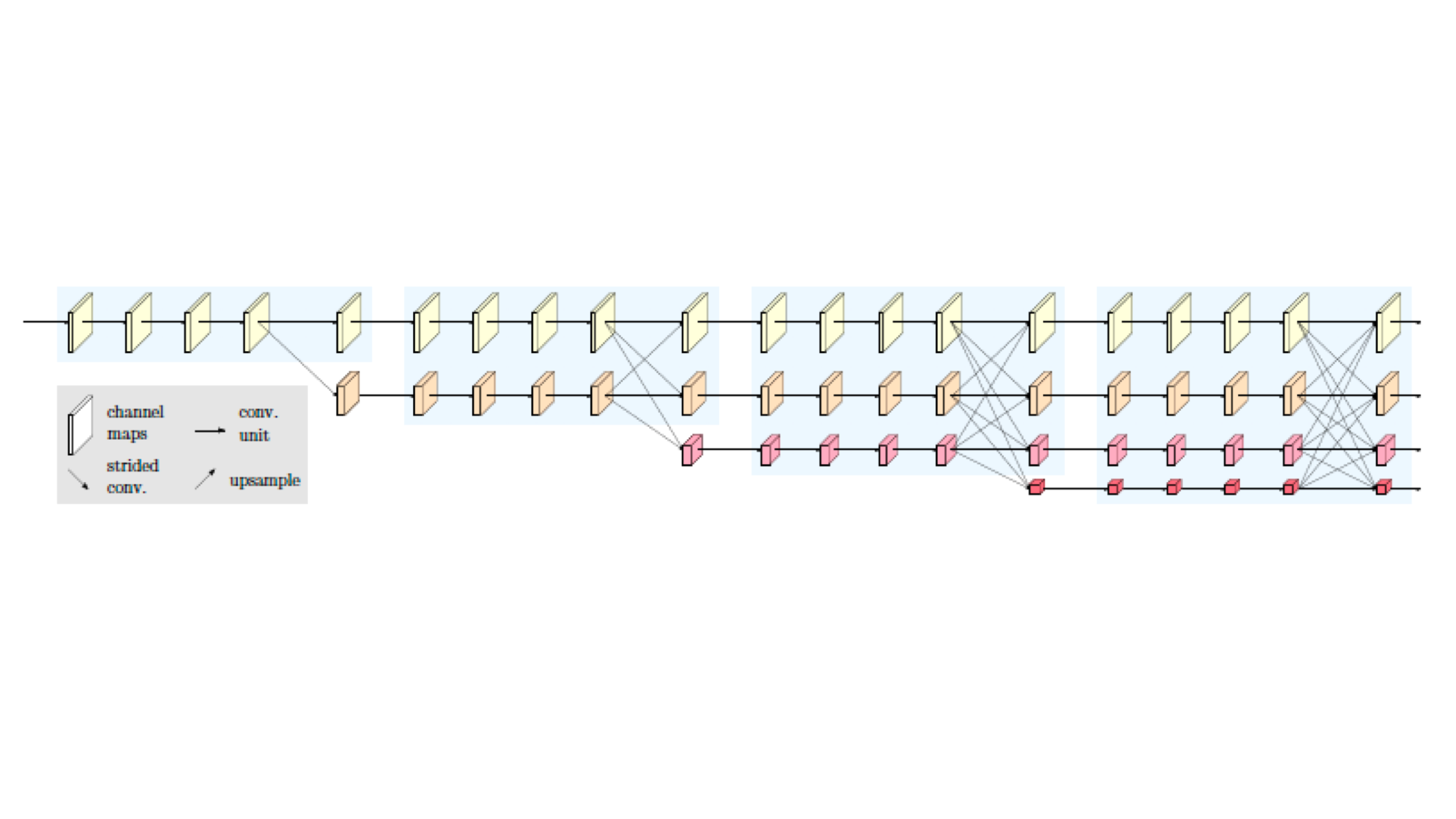}
\caption{{\small{Hrnet image segmentation network architecture}}}
\label{Hrnet}
\end{figure}

\begin{figure}[t]
\begin{centering}
\includegraphics[width=1\linewidth]{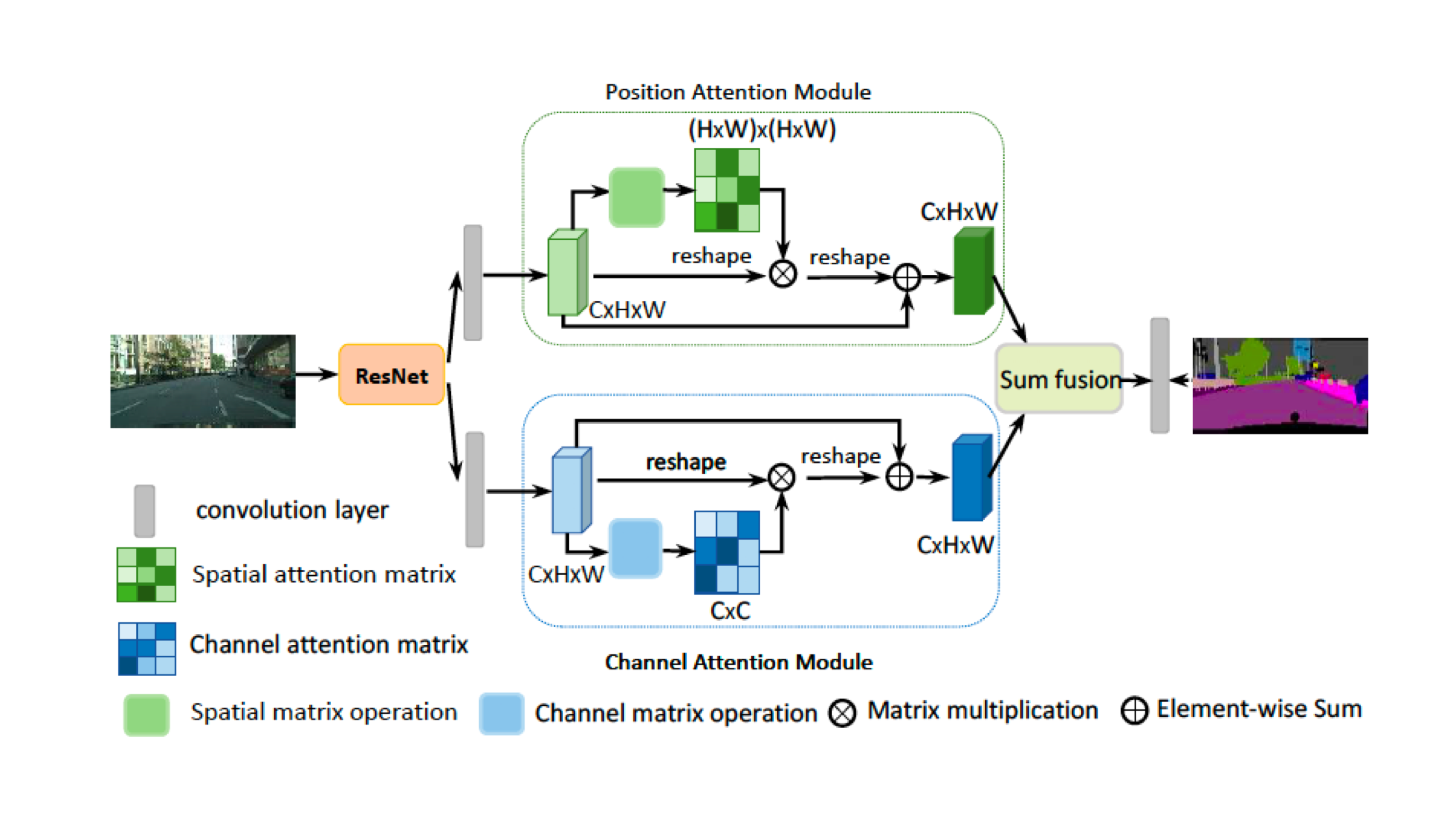}
\par\end{centering}
\caption{{\small{An overview of the Dual Attention Network.}}}
\vspace{-15pt}
\label{Danet}
\end{figure}

\begin{figure}[t]
\begin{centering}
\includegraphics[width=1\linewidth]{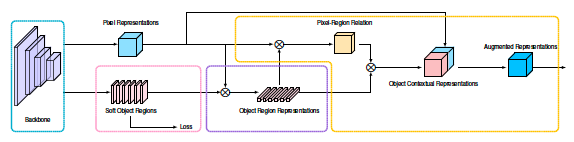}
\par\end{centering}
\caption{{\small{OCRnet Finally, we use the OCRnet network structure to determine the relationship between each object and the pixels on this object, and assign other pixels based on the overall pixel attributes of the object.}}}
\vspace{-15pt}
\label{OCRnet}
\end{figure}

HRNet\cite{wang2020deep} is a particularly well-known network structure recently. In order to maintain high-pixel image characteristics, it uses the connection structure shown in Figure \ref{Hrnet}. The advantage of this structure is that it can ensure the segmentation of the whole large object and retain the low-latitude features and details of image at the same time. According to the network structure of DANet in Figure \ref{Danet}, we use the dual attention mechanism to accurately locate the fourth stage behind HRNet\cite{sun2019deep}. Finally, an OCRnet\cite{yuan2019object} shown in Figure \ref{OCRnet} enables the model to rematch the pixels of object with each pixel in the image.

\subsection{Auxiliary loss}
We use two auxiliary losses to help the final model produce better segmentation results. The first auxiliary loss named $L_{da}$ uses the from of dual attention loss, shown in Figure\ref{Danet}. We use auxiliary loss here to ensure that the result of dual attention module can pay more attention to the overall result and the boundary pixel values of large objects. The second auxiliary loss comes from OCRNet named $L_{ocr}$. This loss is to make the object of OCRNet focus on the object itself. Finally, we use three loss functions for training. The total loss is a weighted sum of three parts, which is defined as following:

\begin{equation}
L_{final}=\alpha{L_{da}}+\beta{L_{ocr}}+\gamma{L_{re}}
\end{equation}

The $\alpha$,$\beta$ and $\gamma$ are auxiliary loss weights, these weights will be introduced in detail in Section \ref{experiment detail}.

\subsection{Data Augmentation}
\label{data augmentation section}
\begin{figure}[t]
\begin{centering}
\includegraphics[width=1\linewidth]{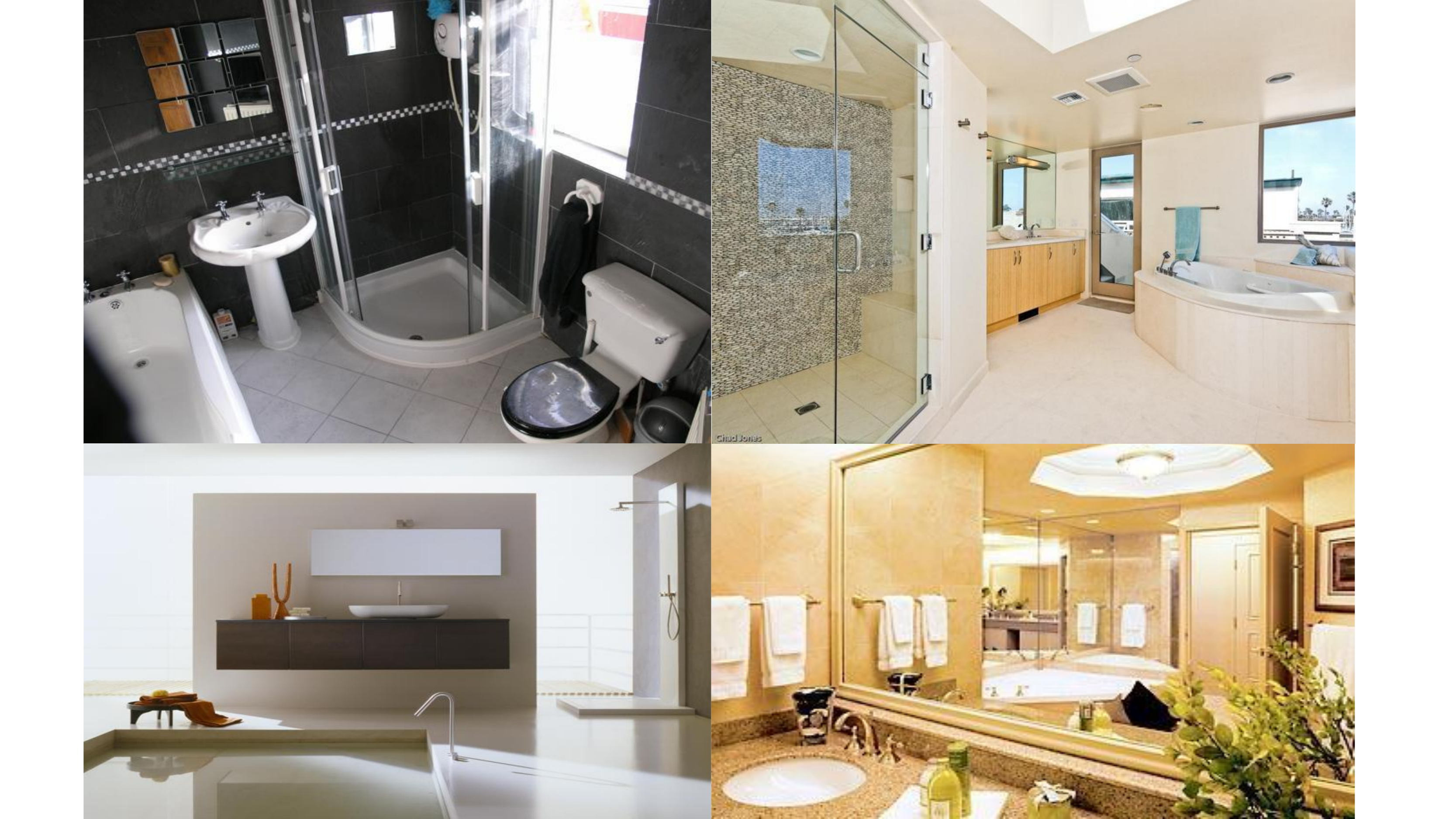}
\par\end{centering}
\caption{{\small{First choose four images and resize them to proper size, then stitch them together according to the position on the map to generate an training image and corresponding label.}}}
\vspace{-15pt}
\label{augmentation}
\end{figure}

There are 150 categories in the ADE20K dataset, and the shape, size and posture of different object in different scenes are different, so it is important to use data augmentation to increase data at different scales. The main method is to use data of different scale during training. four images are superimposed into one image, and at the same time, the corresponding label is also superimposed. This method can make the model pay more attention to the segmentation of small objects. The specific method is shown in Figure \ref{augmentation}

 The purpose of data augmentation is to increase the diversity of the image and the size of the objects in image, so that the corresponding correspondence of the objects in the image becomes smaller, so that the image segmentation algorithm model can segment small objects more robust.

\subsection{Semi-supervised training}
\label{semi-supervised training section}
\begin{figure}[t]
\begin{centering}
\includegraphics[width=1\linewidth]{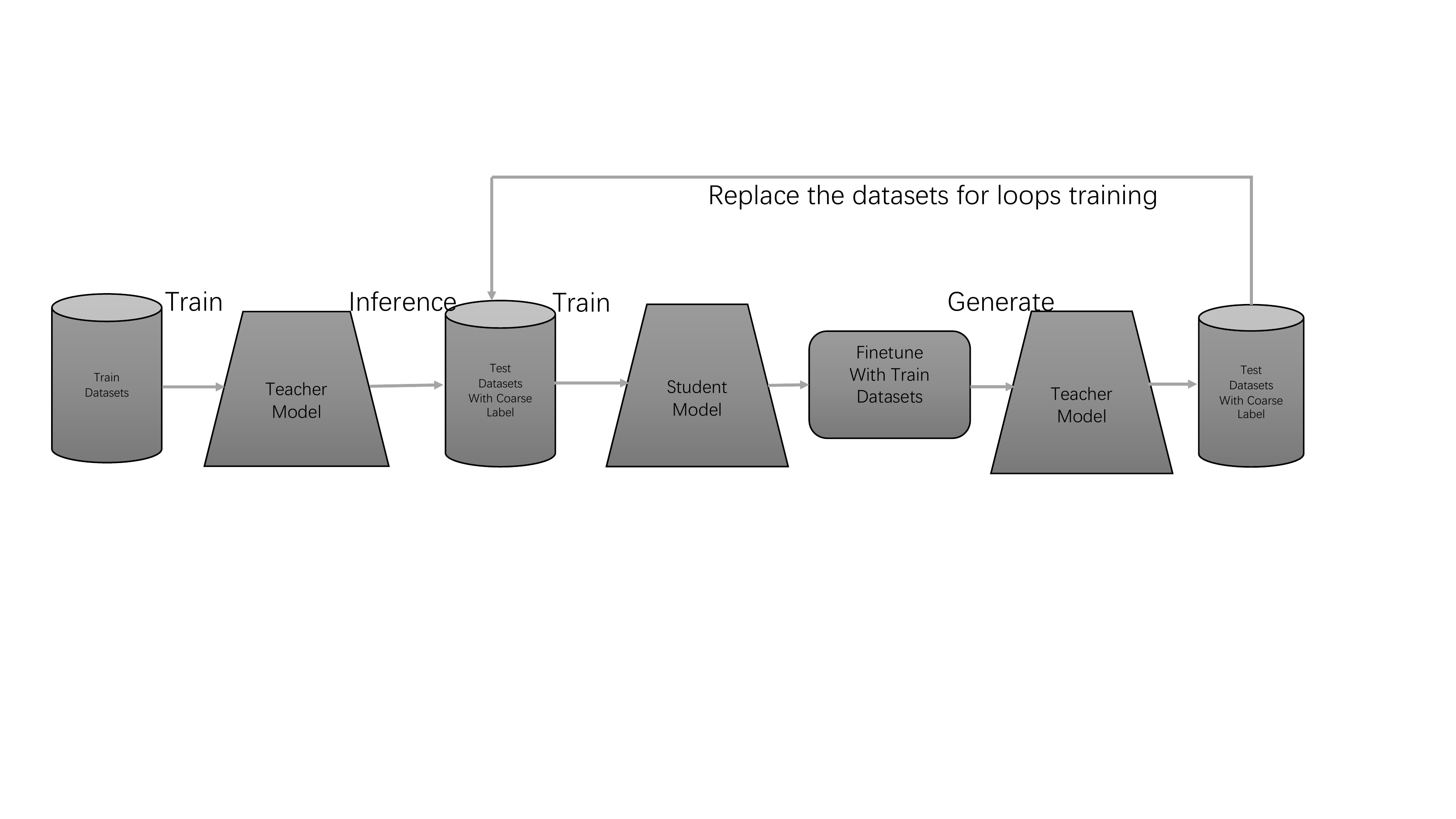}
\par\end{centering}
\caption{{\small{First, train a teacher model using train datasets, and then use the teacher model to inference the test dataset to generate the coarse-grained label. The generated coarse-grained label and the test dataset are paired form the train dataset which is used to train the student model. Then use the train datasets to fintune the student model to generate the teacher model, and then the generated model inference the test dataset, regenerates the new coarse-grained label, Finally useing this formed test dataset pair to replace the first stage test dataset pair and enter the loop training.}}}
\vspace{-15pt}
\label{semi_traing}
\end{figure}

We use a semi-supervised training method\cite{chen2020leveraging} to train the model. The so-called semi-supervised training method refers to the following steps. First, train a basic model as a teacher according to the above methods. Second, predict the test dataset through this teacher, thus forming a test dataset composed of image and coarse-grained label pair. Then, use the image and coarse-grained label pair of this test dataset as the training set of the model to train the student model. Finally, use all the datasets with the test dataset pair to finetune the student model. The student model after training here can be used as a teacher model to complete a second-round semi-supervised learning task. We have used two semi-supervised learning tasks here to achieve state-of-art performance on the test dataset. The whole process is shown in the Figure \ref{semi_traing}

\subsection{Multi-scale training and inference}

The model is sensitive to the scale of data. In order to train a robust model, the training data needs to be multi-scale transformed. Here we take (520, 640) as the basic scale, and the largest scale is (1024, 1280). So multiple scales of data can be used to train the model. The random crop method is also applied to increase the multi-scale information of the data. Similarly, when predicting, we use multiples of the base scale for inference. The multiples are selected as parameters such as 0.5, 1.0, 1.5, 2.0, etc. Finally all inferences are fused to obtain the final result.



\section{Experiments}

\begin{figure}[t]
\begin{centering}
\includegraphics[width=1\linewidth]{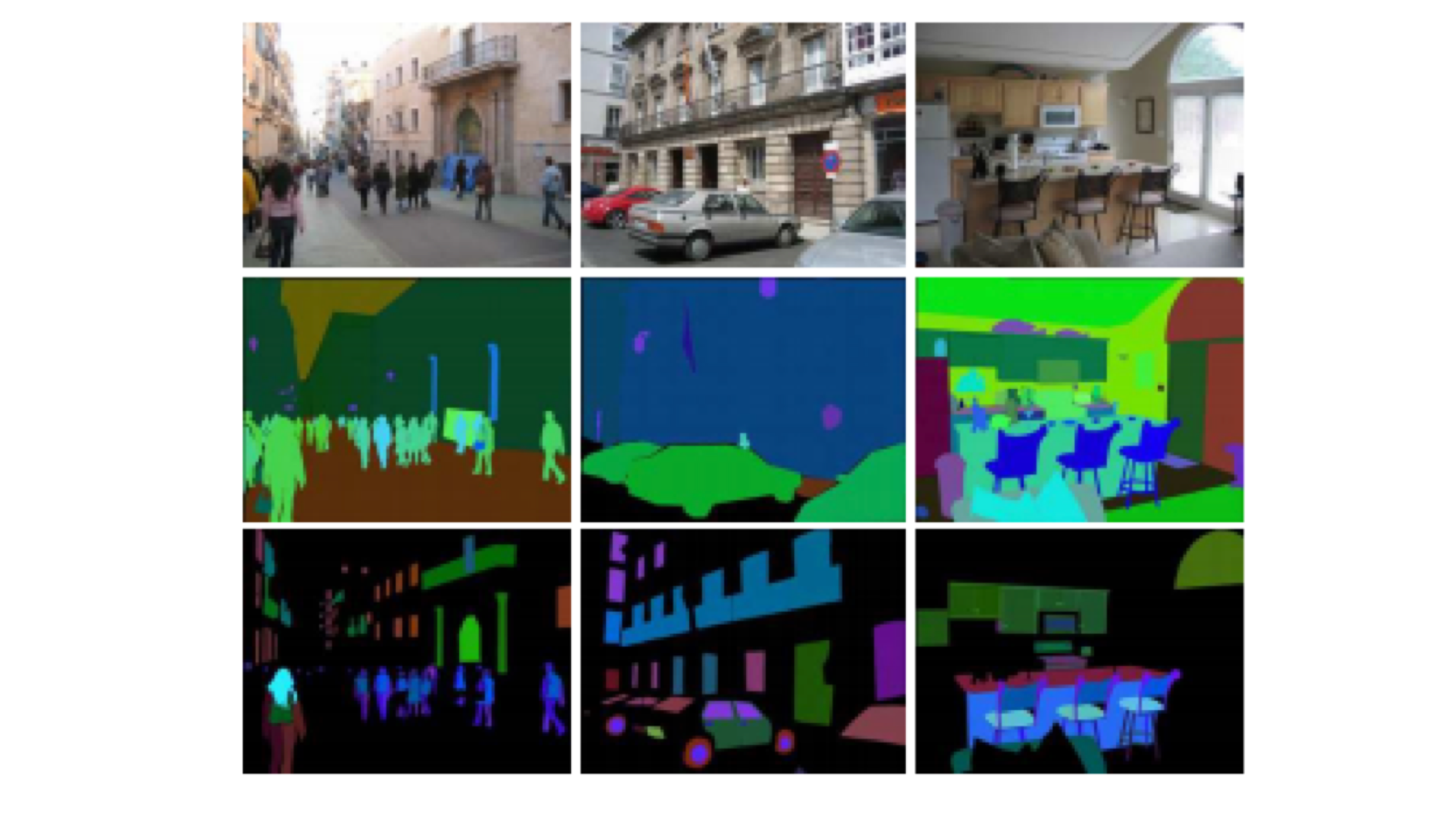}
\par\end{centering}
\caption{{\small{Images in the ADE20K dataset are densely annotated
in details with objects and parts. The first row shows the sample
images, the second row shows the annotation of objects and stuff,
and the third row shows the annotation of object parts.}}}
\vspace{-15pt}
\label{ADE20K}
\end{figure}

The most well-known image segmentation datasets are Pascal Voc\cite{everingham2010pascal}, MSCOCO\cite{lin2014microsoft}, Cityscapes\cite{cordts2016cityscapes},Kitti\cite{geiger2013vision}, ADE20K\cite{zhou2017scene}, etc. We choose the ADE20K datasets to verify the advancedness of our algorithm in the field of panoramic segmentation.
The ADE20K\cite{zhou2017scene} dataset created by MIT which is a very famous panoramic segmentation dataset, which contains 20,000 images for training and 2,000 images for validation, and 3,000 images for test. It has 150 categories in total. The size of different objects varies a lot, and the occlusion between different objects is very serious. At the same time, all the regular objects in the panoramic segmented image are segmented, so the segmentation is very difficult, as shown in Figure \ref{ADE20K}. The final score is composed of two parts, one is the score of pixel accuracy and the other is the score of Miou(Mean IoU). The formula is defined as following:

\begin{equation}
S_{s}=(S_{acc}+S_{Miou})/2.0
\end{equation}

\begin{equation}
Miou=\frac{Intersection(B_{p},B_{gt})}{Union(B_{p},B_{gt})}
\end{equation}

Among them, $B_{p}$ represents the pixel range predicted by the algorithm, and $B_{gt}$ represents ground truth. Intersection() and Union() indicate the intersection and union area between $B_{p}$ and $B_{gt}$.
According to this formula, we can know that pixel-level accuracy and pixel classification of small objects are particularly important to the overall result.

\subsection{Experiment detail}
\label{experiment detail}

At first, we created s new dataset based on the data augmentation method described in \ref{data augmentation section} with a ratio of 1:0.3 data volume. Here we put 6000 images into the training data according to the construction of the training data plus the validation dataset.

Second, we assemble the model according to the above-mentioned way of changing the model. The weights of the auxiliary loss selected here are: 0.1, 0.3, and 1. The weight of 0.1 is the weight of the result after Dual attention named $\alpha$, 0.3 is the first loss weight of OCRNet named$\gamma$, and 1 is the loss of the final generated image named $\gamma$.

Third, we train the model in a semi-supervised way which is described in Section \ref{semi-supervised training section}. First, we train the teacher model on the ADE20K training set. Second, we use the teacher model's prediction as test dataset and get the rough label of this test dataset. Then, we use the test dataset and the rough label as the training set, retrain a new model which is called student model. Finally we use the training dataset to fintune the student model. This forms a training loop. After multiple training loops, we got a model which can achieve state-of-art performance.

Finally, we use multi-scale training and inference. The specific details are to use the three modes of base scale: 520, 640, 800, and then use 7 base multiples when inference: 0.5, 0.75, 1, 1.25, 1.5, 1.75 2.0. All the results are fused together to get the final output result.

\begin{figure}[t]
\begin{centering}
\includegraphics[width=1\linewidth]{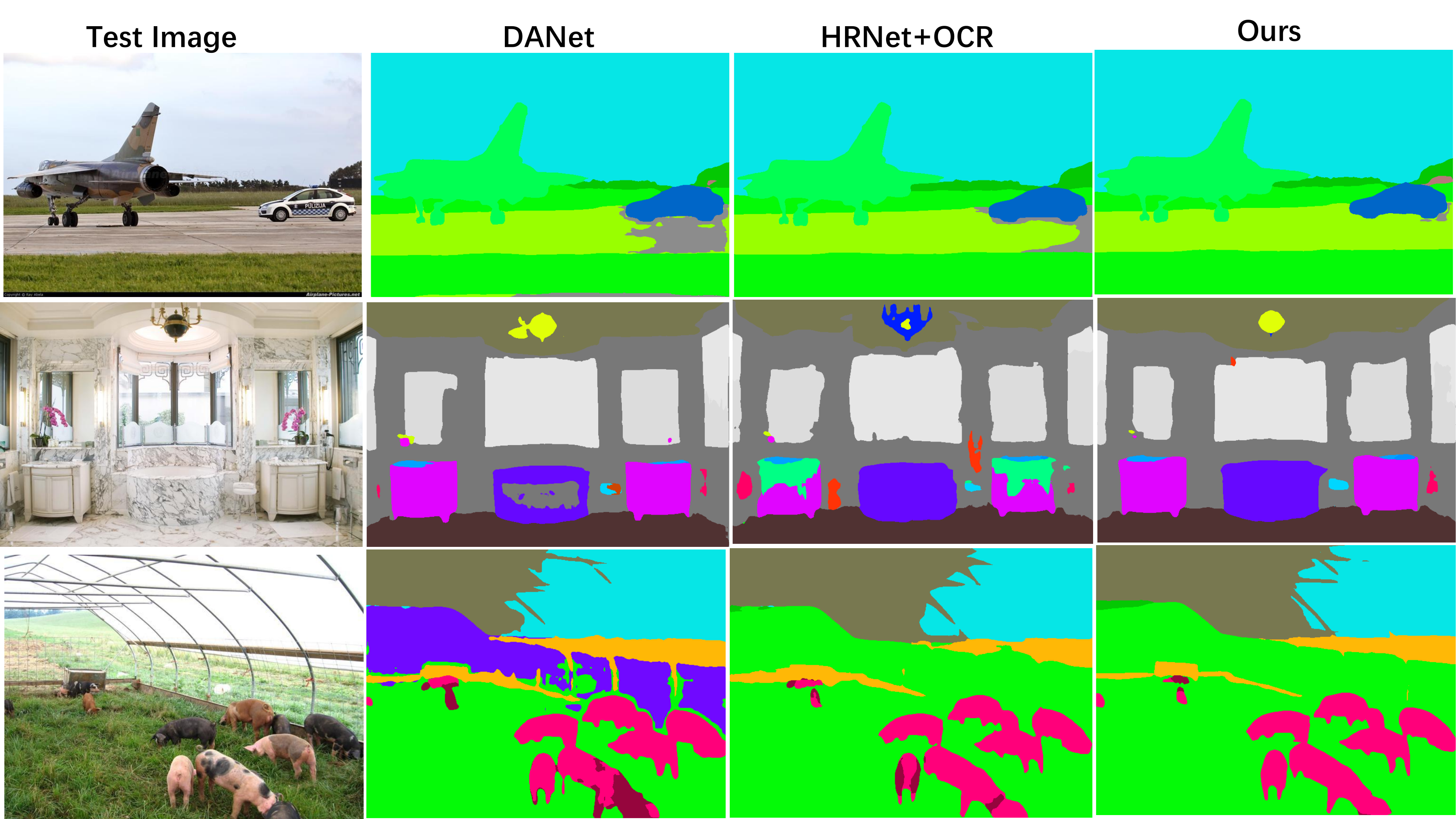}
\par\end{centering}
\caption{{\small{The above figure illustrates that neither HRNet nor DANet can effectively segment the object boundary pixels and little objects, But our algorithm can deal with these well.}}}
\vspace{-15pt}
\label{results}
\end{figure}

Eventually, we can see our results compared with other algorithm in Figure \ref{results}.

\section{Conclusion}

In this paper, we use some recent useful tricks to propose a high-performance algorithm on the famous panoramic segmentation dataset ADE20K. These tricks include: (A) changing the basic network structure to use a multi-stage attention mechanism, which can enable the algorithm to take the pixels of large objects into account, and at the same time can have a good classification effect on boundary pixels of objects with multiple auxiliary loss functions; (B) adopting the way of data enhancement, so that the algorithm can segment small objects better; (C) proposing a  semi-supervised training strategy to make the network more effective; (D) using multi-scale training and inference to obtain state-of-art performance on the dataset ADE20K.

\bibliographystyle{abbrv}
\bibliography{refs}
\end{document}